\documentclass[10pt,twocolumn,letterpaper]{article}

\usepackage{cvpr}      % To produce the REVIEW version
\usepackage{graphicx}
\usepackage{amsmath}
\usepackage{amssymb}
\usepackage{booktabs}
\usepackage{multirow}
\usepackage{color}
\usepackage{colortbl}
\definecolor{LightP}{rgb}{0.95,0.9,1}
\usepackage[pagebackref,breaklinks,colorlinks]{hyperref}

\usepackage[capitalize]{cleveref}
\crefname{section}{Sec.}{Secs.}
\Crefname{section}{Section}{Sections}
\Crefname{table}{Table}{Tables}
\crefname{table}{Tab.}{Tabs.}

\def\cvprPaperID{3116} % *** Enter the CVPR Paper ID here
\def\confName{CVPR}
\def\confYear{2022}

\begin{document}

\title{Virtual Sparse Convolution for Multimodal 3D Object Detection}

\author{Hai Wu$^1$ \quad Chenglu Wen$^{1}$\thanks{Corresponding author} \quad Shaoshuai Shi$^2$ \quad Xin Li$^3$ \quad Cheng Wang$^1$ \\
$^1$Xiamen University \quad $^2$Max-Planck Institute \quad $^3$Texas A\&M University\\
%{\tt\small wuhai@stu.xmu.edu.cn, \{clwen,cwang\}@xmu.edu.cn, shaoshuaics@gmail.com, xinli@tamu.edu }
}

\maketitle
\vspace*{-5mm}

\begin{abstract}
Recently, virtual/pseudo-point-based 3D object detection that seamlessly fuses RGB images and LiDAR data by depth completion has gained great attention. However, virtual points generated from an image are very dense, introducing a huge amount of redundant computation during detection. Meanwhile, noises brought by inaccurate depth completion significantly degrade detection precision. This paper proposes a fast yet effective backbone, termed \textbf{VirConvNet}, based on a new operator \textbf{VirConv} (Virtual Sparse Convolution), for virtual-point-based 3D object detection. VirConv consists of two key designs: (1) \textbf{StVD} (Stochastic Voxel Discard) and (2) \textbf{NRConv} (Noise-Resistant Submanifold Convolution). StVD alleviates the computation problem by discarding large amounts of nearby redundant voxels. NRConv tackles the noise problem by encoding voxel features in both 2D image and 3D LiDAR space. By integrating VirConv, we first develop an efficient pipeline \textbf{VirConv-L} based on an early fusion design. Then, we build a high-precision pipeline \textbf{VirConv-T} based on a transformed refinement scheme. Finally, we develop a semi-supervised pipeline \textbf{VirConv-S} based on a pseudo-label framework. On the KITTI car 3D detection test leaderboard, our VirConv-L achieves \textbf{85\% AP} with a fast running speed of \textbf{56ms}. Our VirConv-T and VirConv-S attains a high-precision of \textbf{86.3\%} and  \textbf{87.2\% AP}, and currently \textbf{rank 2nd and 1st}\footnote[1]{On the date of CVPR deadline,~\ie, Nov.11, 2022}, respectively. The code is available at \url{https://github.com/hailanyi/VirConv}.
\end{abstract}
\section{Introduction}
\label{sec:intro}
\begin{figure}[t]
  \centering
   \includegraphics[width=1\linewidth]{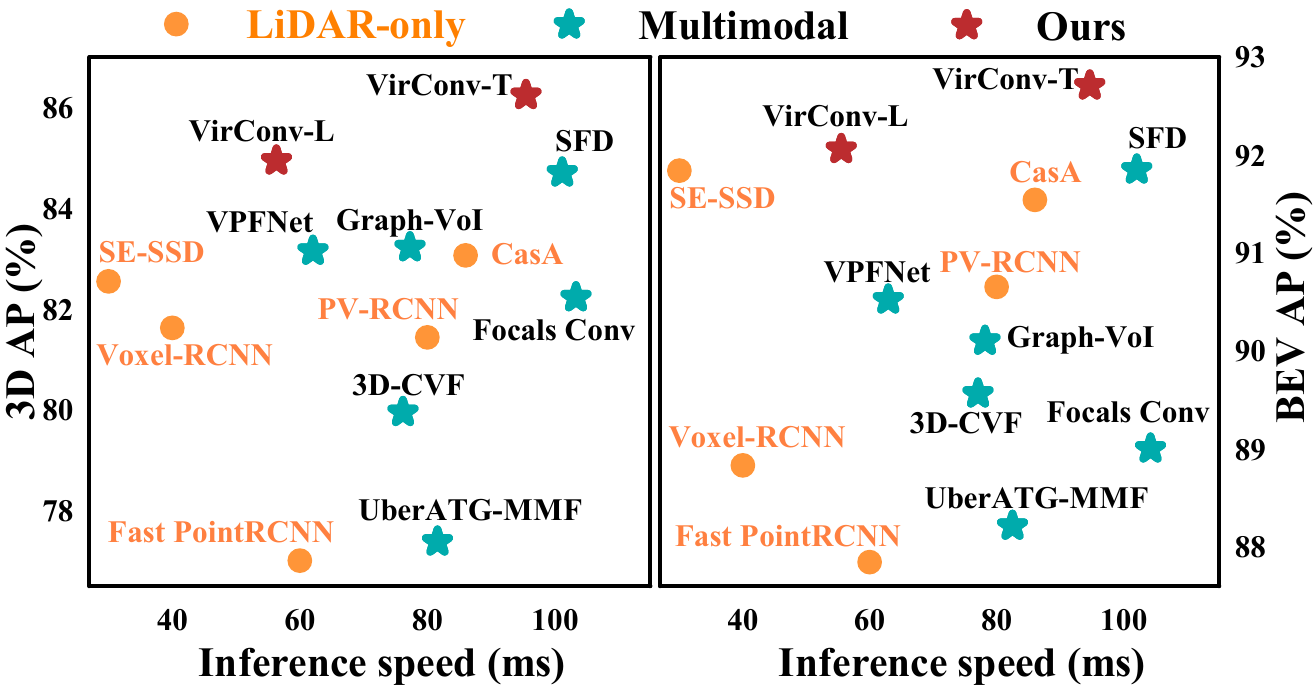}
   \caption{Our VirConv-T achieves top average precision (AP) on both 3D and BEV moderate car detection in the KITTI benchmark (more details are in Table~\ref{tab:kitti_test_ap}). Our VirConv-L runs fast at 56ms with competitive AP.}
   \label{fig:performance}
\end{figure}

3D object detection plays a critical role in autonomous driving~\cite{MPVN,MVP}. The LiDAR sensor measures the depth of scene~\cite{MV3D} in the form of a point cloud and enables reliable localization of objects in various lighting environments. While LiDAR-based 3D object detection has made rapid progress in recent years~\cite{PartA2,3DIoULoss,PointPillars,STD,3DSSD,PointRCNN,CT3D,Pyramid}, its performance drops significantly on distant objects, which inevitably have sparse sampling density in the scans. 
Unlike LiDAR scans, color image sensors provide high-resolution sampling and rich context data of the scene. 
The RGB image and LiDAR data can complement each other and usually boost 3D detection performance~\cite{F-PointNet,UberATG-MMF,BEVFusion,TransFusion,FConv}.

Early methods~\cite{MVX-Net, Pointaugmenting, PointPainting} extended the features of LiDAR points with image features, such as semantic mask and 2D CNN features. They did not increase the number of points; thus, the distant points still remain sparse. In contrast, the methods based on virtual/pseudo points (for simplicity, both denoted as \textbf{virtual points} in the following) enrich the sparse points by creating additional points around the LiDAR points. For example, MVP~\cite{MVP} creates the virtual points by completing the depth of 2D instance points from the nearest 3D points. SFD~\cite{SFD} creates the virtual points based on depth completion networks~\cite{PENet}. The virtual points complete the geometry of distant objects, showing the great potential for high-performance 3D detection.

\begin{figure}[t]
  \centering
   \includegraphics[width=1\linewidth]{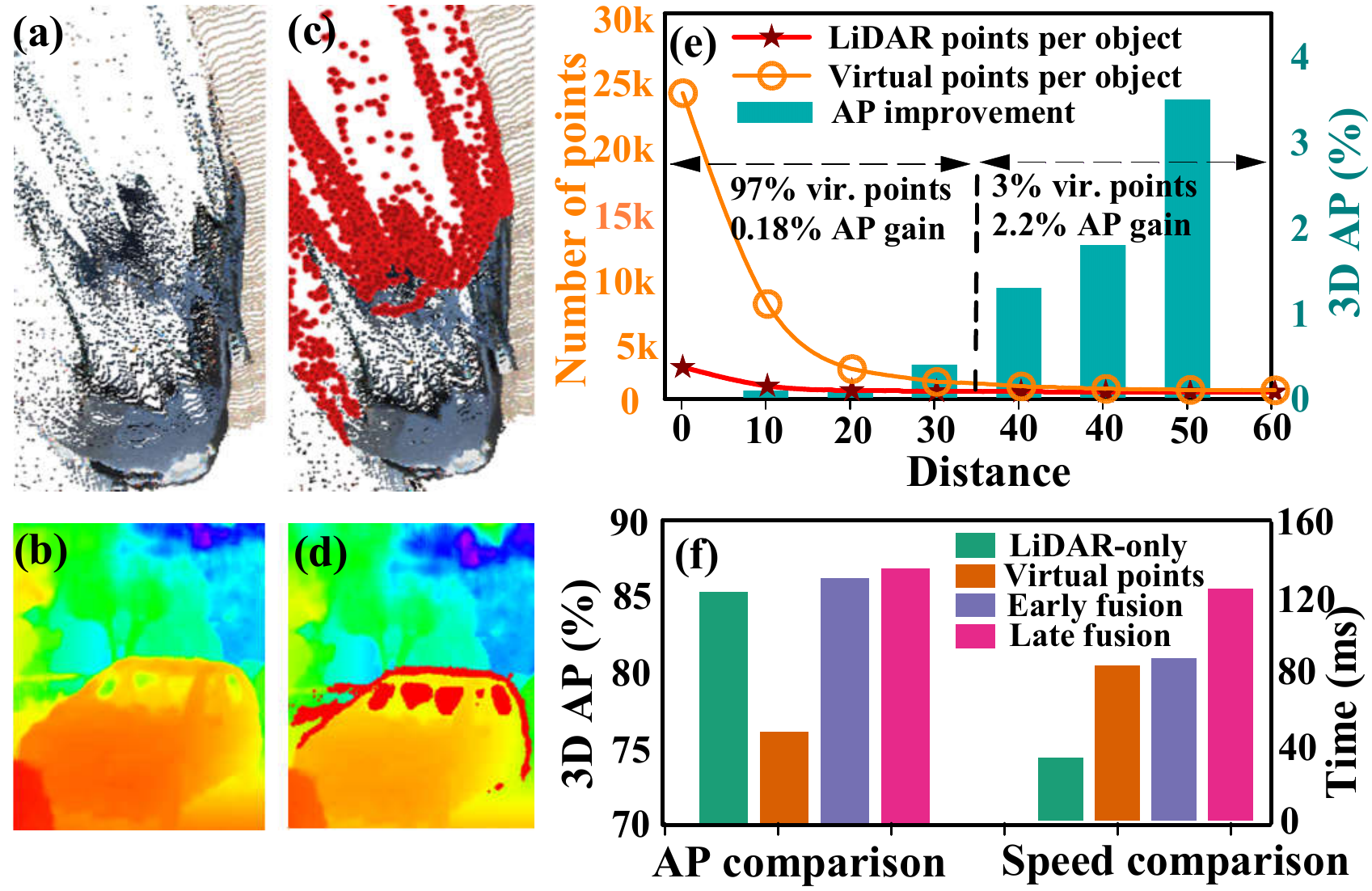}
   \caption{The noise problem and density problem of virtual points.  (a) Virtual points in 3D space. (b) Virtual points in 2D space. (c) Noises (red) in 3D space. (d) Noises (red) distributed on 2D instance boundaries. (e) Virtual points number versus AP improvement along different distances by using Voxel-RCNN~\cite{Voxel-RCNN} with late fusion (details see Sec.~\ref{sec:vpdf}). (f) Car 3D AP and inference time using Voxel-RCNN~\cite{Voxel-RCNN} with LiDAR-only, virtual points-only, early fusion, and late fusion (details see Sec.~\ref{sec:vpdf}), respectively.}
   \label{fig:motivation}
\end{figure}

However, virtual points generated from an image are generally very dense. Taking the KITTI~\cite{KITTI} dataset as an example, an 1242$\times$375 image generates 466k virtual points (\textbf{$\sim$27$\times$} more than the LiDAR scan points). This brings a huge computational burden and causes a severe efficiency issue (see Fig.~\ref{fig:motivation} (f)). Previous work addresses the density problem by using a larger voxel size~\cite{PointPillars, CenterPoint} or by randomly down-sampling~\cite{RandLA-Net} the points. However, applying such methods to virtual points will inevitably sacrifice useful shape cues from faraway points and result in decreased detection accuracy.

\begin{figure*}[t]
  \centering
   \includegraphics[width=0.95\linewidth]{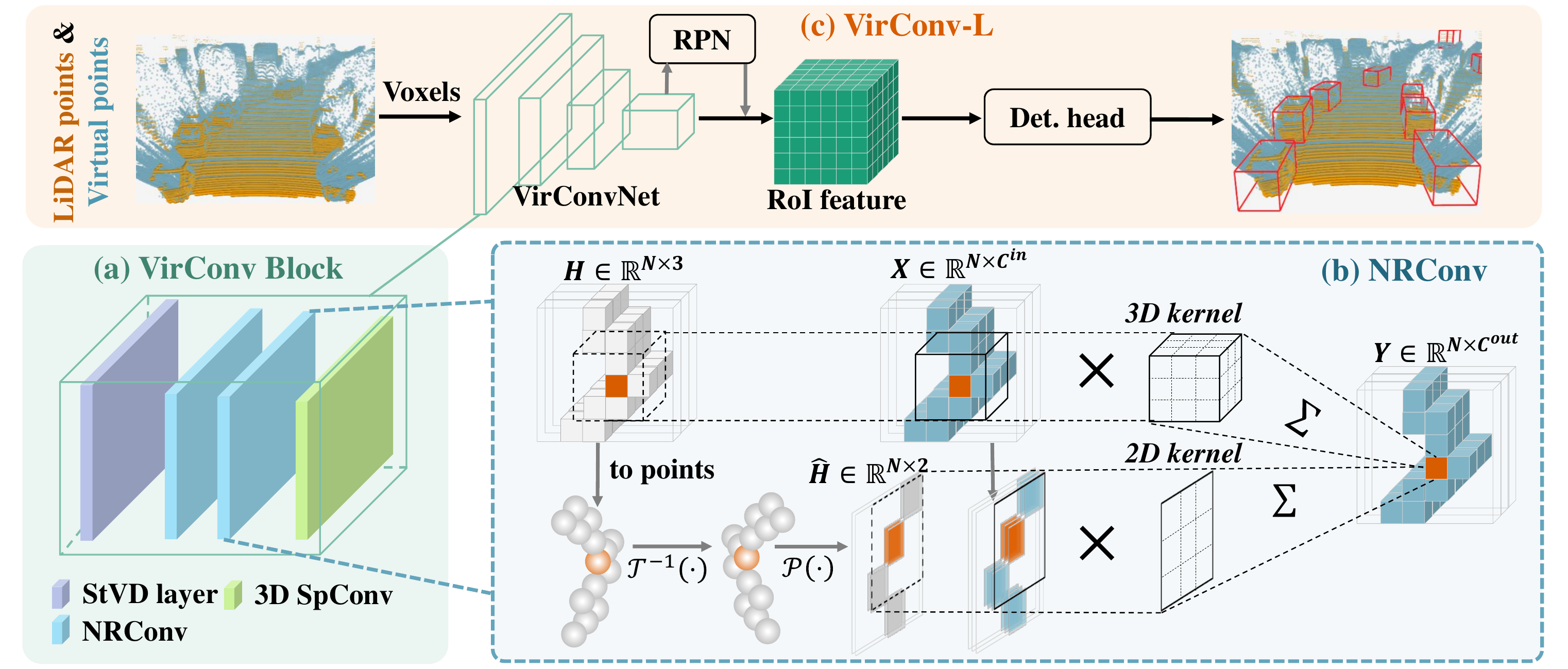}
   \caption{(a) VirConv block consists of a StVD layer, some NRConv layers and a 3D SpConv layer. (b) NRConv projects the voxels back to image space, and encodes virtual point features in both 2D and 3D space. (c) VirConv-L fuses the LiDAR points and the virtual points into a single point cloud, and encodes the multimodal features by our VirConvNet for 3D detection.  }
   \label{fig:framework}
\end{figure*}

Another issue is that the depth completion can be inaccurate, and it brings a large amount of noise in the virtual points (see Fig.~\ref{fig:motivation} (c)).
Since it is very difficult to distinguish the noises from the background in 3D space, the localization precision of 3D detection is greatly degraded. 
In addition, the noisy points are non-Gaussian distributed, and can not be filtered by conventional denoising algorithms~\cite{BMD,Reviewfilter}. 
Although recent semantic segmentation network~\cite{CNNDenoising} show promising results, they generally require extra annotations.

To address these issues, this paper proposes a VirConvNet pipeline based on a new Virtual Sparse Convolution (VirConv) operator. 
Our design builds on \textbf{two main observations}.
(1) First, geometries of nearby objects are often relatively complete in LiDAR scans. Hence, most virtual points of nearby objects only bring marginal performance gain (see Fig.~\ref{fig:motivation} (e)(f)), but increase the computational cost significantly.  
(2) Second, noisy points introduced by inaccurate depth completions are mostly distributed on the instance boundaries (see Fig.~\ref{fig:motivation} (d)). 
They can be recognized in 2D images after being projected onto the image plane. 

Based on these two observations, we design a \textbf{StVD} (Stochastic Voxel Discard) scheme to retain those most important virtual points by a bin-based sampling, namely, discarding a huge number of nearby voxels while retaining faraway voxels. This can greatly speed up the network computation. 
We also design a \textbf{NRConv} (Noise-Resistant Submanifold Convolution) layer to encode geometry features of voxels in both 3D space and 2D image space. 
The extended receptive field in 2D space allows our NRConv to distinguish the noise pattern on the instance boundaries in 2D image space. Consequently, the negative impact of noise can be suppressed. 

We develop three multimodal detectors to demonstrate the superiority of our VirConv: (1) a lightweight \textbf{VirConv-L} constructed from Voxel-RCNN~\cite{Voxel-RCNN}; (2) a high-precision \textbf{VirConv-T} based on multi-stage~
\cite{CasA} and multi-transformation~\cite{TED} design; (3) a semi-supervised \textbf{VirConv-S} based on a pseudo-label~\cite{3DIoUMatch} framework. The effectiveness of our design is verified by extensive experiments on the widely used KITTI dataset~\cite{KITTI} and nuScenes dataset~\cite{nuScenes}.
Our contributions are summarized as follows: 

\begin{itemize}
\item We propose a \textbf{VirConv} operator, which effectively encodes voxel features of virtual points by \textbf{StVD} and \textbf{NRConv}. The StVD discards a huge number of redundant voxels and substantially speeds up the 3D detection prominently. The NRConv extends the receptive field of 3D sparse convolution to the 2D image space and significantly reduces the impact of noisy points. 

\item Built upon VirConv, we present three new multimodal detectors: a \textbf{VirConv-L}, a \textbf{VirConv-T}, and a semi-supervised \textbf{VirConv-S} for efficient, high-precision, and semi-supervised 3D detection, respectively.

\item Extensive experiments demonstrated the effectiveness of our design (see Fig.~\ref{fig:performance}). On the KITTI leaderboard, our VirConv-T and VirConv-S currently \textbf{rank 2nd and 1st}, respectively. Our VirConv-L runs at \textbf{56ms} with competitive precision.
\end{itemize}
\section{Related Work}

 \textbf{LiDAR-based 3D object detection.}
LiDAR-based 3D object detection has been widely studied in recent years. Early methods project the point clouds into a 2D Bird's eye view (BEV) or range view images~\cite{MV3D,Birdnet} for 3D detection. Recently, voxel-based sparse convolution~\cite{SECOND,PointPillars,Voxel-RCNN,SA-SSD} and point-based set abstraction~\cite{PointRCNN,STD,3DSSD,PV-RCNN} have become popular in designing effective detection frameworks. 
However, the scanning resolution of LiDAR is generally very low for distant objects. The LiDAR-only detectors usually suffer from such sparsity. This paper addresses this problem by introducing RGB image data in a form of virtual points.

 \textbf{Multimodal 3D object detection.}
The RGB image and LiDAR data can complement each other and usually boost 3D detection performance. Early methods extend the features of LiDAR points with image features~\cite{MVX-Net, Pointaugmenting, PointPainting}. Some works encode the feature of two modalities independently and fuse the two features in the local Region of Interest (RoI)~\cite{AVOD-FPN,FUTR3D} or BEV plane~\cite{BEVFusion}. We follow the recent work that fuses the two data via virtual points~\cite{MVP,SFD}. The virtual points explicitly complete the geometry of distant objects by depth estimation, showing the great potential for high-performance 3D detection. But virtual points are extremely dense and often noisy. This paper addresses these problems through two new schemes, StVD and NRConv, respectively.

\textbf{3D object detection with re-sampled point clouds.}
The points captured by LiDAR are generally dense and unevenly distributed. Previous work speeds up the network by using a larger voxel size~\cite{PointPillars, CenterPoint} or by randomly down-sampling~\cite{RandLA-Net} the point clouds. However, applying these methods to the virtual points will significantly decrease the useful geometry cues, especially for the faraway objects. Different from that, our StVD retains all the useful faraway voxels and speeds up the network by discarding nearby redundant voxels. 

\textbf{Noise handling in 3D vision.}
Traditional methods handle the noises by filtering algorithm~\cite{BMD,Reviewfilter,GuidedFilter}. Recently, score-based~\cite{SBDenoising} and semantic segmentation networks~\cite{CNNDenoising} are developed for point cloud noise removal. Different from the traditional noises that are randomly distributed in 3D space, the noises brought by inaccurate depth completion are mostly distributed on 2D instance boundaries. Although the noise can be roughly removed by some 2D edge detection method~\cite{BiEdge}, this will sacrifice the useful boundary
points of object. We design a new scheme, NRConv, that extends the receptive field of 3D sparse convolution to 2D image space, distinguishing the noise pattern without the loss of useful boundary points.

\textbf{Semi-supervised 3D object detection.}
Recent semi-supervised methods boost 3D object detection by a large amount of unlabeled data. Inspired by the pseudo-label-based framework~\cite{3DIoUMatch,SESS,SS3D}, we also constructed a VirConv-S pipeline to perform semi-supervised multimodal 3D object detection. 

\section{VirConv for Multimodal 3D Detection}
\label{sec:virpoint}

This paper proposes VirConvNet, based on a new VirConv operator, for virtual-point-based multimodal 3D object detection. 
As shown in Fig.~\ref{fig:framework},  VirConvNet first converts points into voxels, and gradually encodes voxels into feature volumes by a series of VirConv block with $1\times$, $2\times$, $4\times$ and $8\times$ downsampling strides. The VirConv block consists of three parts (see Fig.~\ref{fig:framework} (a)): (1) an StVD layer for speeding up the network and improving density robustness; (2) multiple NRConv layers for encoding features and decreasing the impact of noise; (3) a 3D SpConv layer for down-sampling the feature map. Based on the VirConv operator, we build three detectors for efficient, accurate, and semi-supervised multimodal 3D detection, respectively.

\subsection{Virtual Points for Data Fusion}
\label{sec:vpdf}

Many recent 3D detectors use virtual points~\cite{MVP} (pseudo points~\cite{SFD}) generated from an image by depth completion algorithms to fuse RGB and LiDAR data. 
We denote the LiDAR points and virtual points as $\mathbf{P}$ and $\mathbf{V}$, respectively. 
Recently, two popular fusion schemes have been applied for 3D object detection: 
(1) early fusion~\cite{MVP}, which fuses $\mathbf{P}$ and $\mathbf{V}$ into a single point cloud $\mathbf{P}^*$ and performs 3D object detection using existing detectors, and 
(2) late fusion~\cite{SFD}, which encodes the features of $\mathbf{P}$ and $\mathbf{V}$ by different backbone networks and fuses the two types of features in BEV plane or local RoI.
%\textcolor{red}{(``late fusion'' reads very awkward. Is this an commonly used term or a self-invented word?)}
However, both fusion methods suffer from the dense and noisy nature of virtual points. 

\textbf{(1) Density problem.}
As motioned in Section~\ref{sec:intro}, the virtual points are usually very dense. They introduce a huge computational burden, which significantly decreases the detection speed (e.g., more than \textbf{2$\times$} in Fig.~\ref{fig:motivation} (f)). 
Existing work tackles the density issue by using a larger voxel size~\cite{PointPillars} or by randomly down-sampling~\cite{RandLA-Net} the points. But these methods will inevitably sacrifice the shape cues from the virtual points, especially for the faraway object. Based on a pilot experiment on the KITTI dataset~\cite{KITTI} using the Voxel-RCNN~\cite{Voxel-RCNN} with a late fusion, we observed that a huge number of virtual points introduced for nearby objects are redundant. Specifically, \textbf{97\%} of virtual points from the nearby objects bring only a \textbf{0.18\%} performance improvement, while \textbf{3\%} of virtual points for the faraway objects bring a \textbf{2.2\%} performance improvement. 
The reason is that the geometry of nearby objects
is relatively complete for LiDAR points. 
Such virtual points generally bring marginal performance gain but increase unnecessary computation. Motivated by this observation, we design an StVD (Stochastic Voxel Discard) scheme, which alleviates the computation problem by discarding nearby redundant voxels. In addition, the points of the distant object are much sparser than the nearby objects (see Fig.~\ref{fig:motivation} (e)). The StVD can simulate sparser training samples to improve detection robustness.

\textbf{(2) Noise problem.}
The virtual points generated by the depth completion network are usually noisy. An example is shown in Fig.~\ref{fig:motivation} (c). The noise is mostly introduced by the inaccurate depth completion, and is hardly distinguishable in  3D space. By using only virtual points, the detection performance drops $\sim$9\% AP compared with the LiDAR-only detector (see Fig.~\ref{fig:motivation} (f)). In addition, the noisy points are non-Gaussian distributed, and cannot be filtered by traditional denoising algorithms~\cite{BMD,Reviewfilter}. 
We observed that noise is mainly distributed on the instance boundaries (see Fig.~\ref{fig:motivation} (d)) and can be more easily recognized in 2D images. Although the edge detection~\cite{BiEdge} could be applied here to roughly remove the noise, this will sacrifice the useful boundary points which are beneficial to the object's shape and position estimation. Our idea is to extend the receptive field of sparse convolution to the 2D image space, and distinguish the noise without the loss of shape cues. 

\subsection{Stochastic Voxel Discard}
\label{sec:svd}

To alleviate the computation problem and improve the density robustness for the virtual-point-based detector, we develop the StVD. It consists of two parts: (1) input StVD, which speeds up the network by discarding input voxels of virtual points during both the training and inference process; (2) layer StVD, which improves the density robustness by discarding voxels of virtual points at every VirConv block during only the training process.

\begin{figure}[t]
  \centering
   \includegraphics[width=1\linewidth]{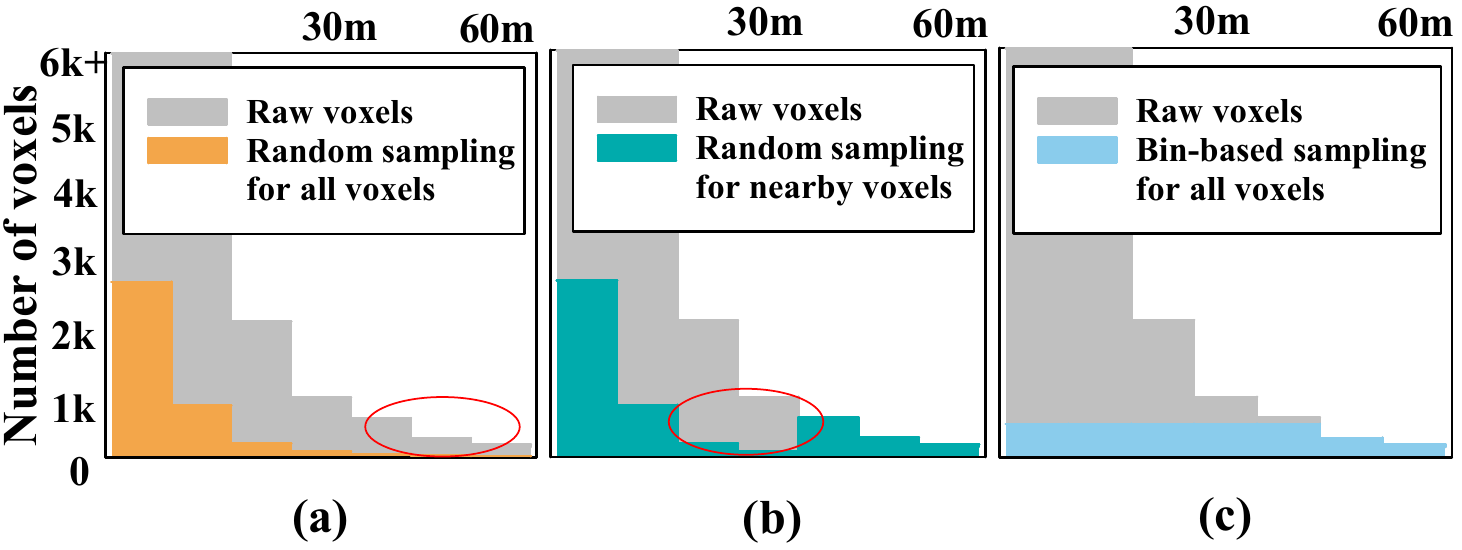}
   \caption{(a)(b) show the voxel distributions after random sampling for all and nearby voxels, respectively. (c) shows the voxel distribution after bin-based sampling for all voxels.}
   \label{fig:method_sampling}
\end{figure}

\textbf{Input StVD.}
Two naive methods can keep less input voxels: (1) random sampling or (2) farthest point sampling (FPS).
However, the random sampling usually keeps unbalanced voxels at different distances and inevitably sacrifices some useful shape cues (in the red region at Fig.~\ref{fig:method_sampling} (a) (b)). In addition, FPS needs huge extra computation when down-sampling the huge number of virtual points due to the high computational complexity ($O(n^2)$).
To tackle this problem, we introduce a bin-based sampling strategy to perform efficient and balanced sampling (see Fig.~\ref{fig:method_sampling} (c)). 
Specifically, We first divide the input voxels into $N^b$ bins (we adopt $N^b=10$ in this paper) according to different distances. For the nearby bins ($\leq$30m based on the statistics in Fig.~\ref{fig:motivation} (e)), we randomly keep a fixed number ($\sim$ 1K) of voxels. For distant bins, we keep all of the inside voxels. After the bin-based sampling, we discard about \textbf{90\%} (which achieves the best precision-efficiency trade-off, see Fig.~\ref{tab:ablation_svd_rate}) of redundant voxels and it speeds up the network by about \textbf{2 times}.

\textbf{Layer StVD.}
To improve the robustness of detection from sparse points, we also develop a layer StVD which is applied to the training process.
Specifically, we discard voxels at each VirConv block to simulate sparser training samples. We adopt a discarding rate of 15\% in this paper (the layer StVD rate is discussed in Fig.~\ref{tab:ablation_svd_rate}). The layer StVD serves as a data augmentation strategy to help enhance the 3D detector's training.

\subsection{Noise-Resistant Submanifold Convolution}
As analyzed in Section~\ref{sec:vpdf}, the noise introduced by the inaccurate depth completion can hardly be recognized from 3D space but can be easily recognized from 2D images. We develop an NRConv (see Fig.~\ref{fig:framework} (b)) from the widely used submanifold sparse convolution~\cite{SpConv} to address the noise problem. Specifically, given $N$ input voxels formulated by a 3D indices vector
$\mathbf{H}\in \mathbb{R}^{N\times 3}$ and a features vector
$\mathbf{X}\in \mathbb{R}^{N\times C^{in}}$, we encode the noise-resistant geometry features $\mathbf{Y}\in \mathbb{R}^{N\times C^{out}}$ in both 3D and 2D image space, where $C^{in}$ and $C^{out}$ denote the number of input and output feature channels respectively.

\textbf{Encoding geometry features in 3D space.}
For each voxel feature $ X_i$ in $\mathbf{X}$, we first encode the geometry features by the 3D submanifold convolution kernel $\mathcal{K}^{3D}(\cdot)$. Specifically, the geometry features $\hat{X}_i\in\mathbb{R}^{C^{out}/2}$ are calculated from the non-empty voxels within $3\times3\times3$ neighborhood based on the corresponding 3D indices as

\vspace{-4mm}
\begin{align} 
\hat{X}_i = \mathcal{R}\left\{\mathcal{K}^{3D} \left(  X_i, X_i^{(f_{1})}, ..., X_i^{(f_{j})} \right)\right\}, %\text{ for } i = 1, ..., N,
\end{align}
\vspace{-4mm}

\noindent where $X_i^{(f_{1})}, ..., X_i^{(f_{j})}$ denote neighbor features generated by $\mathbf{H}$, and $\mathcal{R}$ denotes the nonlinear activation function.

 \textbf{Encoding noise-aware features in 2D image space.}
The noise brought by the inaccurate depth completion significantly degrade the detection performance. Since the noise is mostly distributed on the 2D instance boundaries, we extend the convolution receptive field to the 2D image space and encode the noise-aware features using the 2D neighbor voxels. Specifically, we first convert the 3D indices to a set of grid points based on the voxelization parameters (the conversion denoted as $\mathcal{G}(\cdot)$). Since state-of-the-art detectors~\cite{SFD, Voxel-RCNN} also adopt the transformation augmentations (the augmentation denoted as $\mathcal{T}(\cdot)$) such as rotation and scaling, the grid points are generally misaligned with the corresponding image. Therefore, we transform the grid points backward into the original coordinate system based on the data augmentation parameters. Then we project the grid points into the 2D image plane based on the LiDAR-Camera calibration parameters (with the projection denoted as $\mathcal{P}(\cdot)$). The overall projection can be summarized by 

\vspace{-4mm}
\begin{align}
    \mathbf{\hat{H}} = \mathcal{P} \left(\mathcal{T}^{-1}\left(\mathcal{G}\left(\mathbf{H}\right)\right)\right),
\end{align}
\vspace{-4mm}

\noindent where $\mathbf{\hat{H}}\in\mathbb{R}^{N\times2}$ denotes the 2D indices vector.
For each voxel feature $X_i\in \mathbb{R}^{C^{in}}$, we then calculate the noise-aware features $\tilde{X}_i\in\mathbb{R}^{C^{out}/2}$ from the non-empty voxels within a $3\times3$ neighborhood based on the corresponding 2D indices. 

\vspace{-4mm}
\begin{align} 
\tilde{X}_i = \mathcal{R}\left\{\mathcal{K}^{2D} \left (  X_i, \tilde{X}_i^{(f_{1})}, ..., \tilde{X}_i^{(f_{k})} \right) \right\}, %\text{ for } i = 1, ..., N,
\end{align}
\vspace{-4mm}

\noindent where $\tilde{X}_i^{(f_{1})}, ..., \tilde{X}_i^{(f_{k})}$ denote the neighbor voxel features generated by $\mathbf{\hat{H}}$, and  $\mathcal{K}^{2D}(\cdot)$ denote the 2D submanifold convolution kernel. If there are multiple features in a single 2D neighbor voxel, we perform max-pooling and keep one feature in each voxel to perform the 2D convolution. 

After the 3D and 2D features encoding, we adopt a simple concatenation to implicitly learn a noise-resistant feature. 
Specifically, we finally concatenate $\hat{X}_i$ and $\tilde{X}_i$ to obtain the noise-resistant feature vector $\mathbf{Y}\in\mathbb{R}^{N\times C^{out}}$ as
\begin{equation}
   \mathbf{Y} =\left[\left[\hat{X}_i,\tilde{X}_i\right]^T,...,\left[\hat{X}_N,\tilde{X}_N\right]^T \right]^T.
\end{equation}
Different from related noise segmentation and removal~\cite{CNNDenoising} methods, our NRConv implicitly distinguishes the noise pattern by extending the receptive field to 2D image space. Consequently, the impact of noise is suppressed without lose of shape cues.
%-------------------------------------------------------------------------

\begin{figure}[t]
  \centering
   \includegraphics[width=1\linewidth]{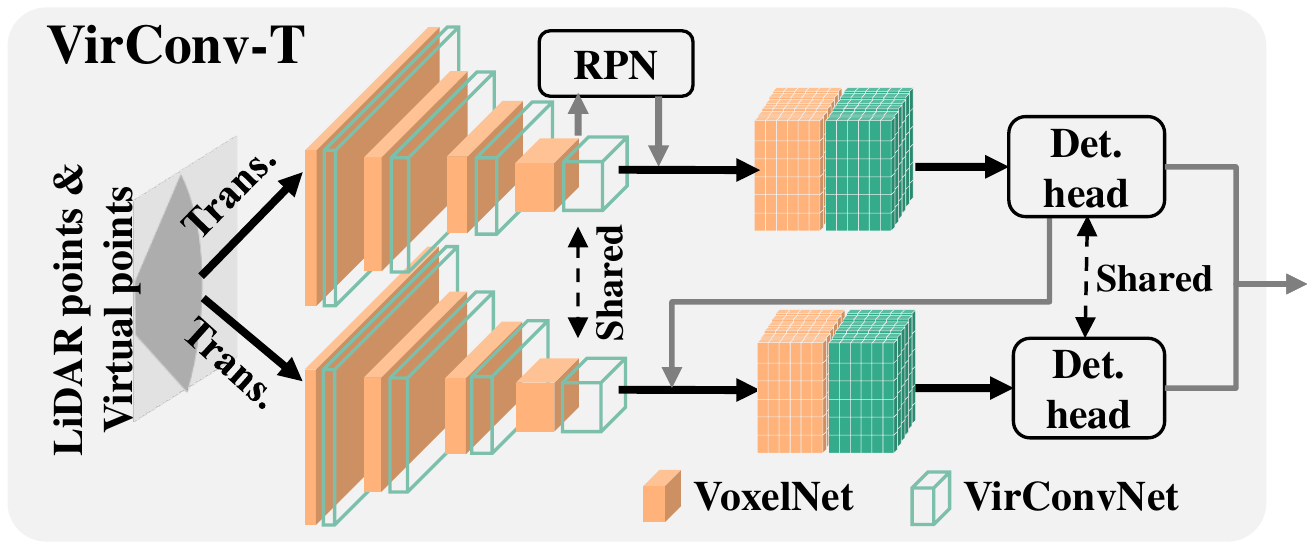}
   \caption{Transformed refinement scheme. The inputs are first transformed with different rotations and reflections. Then, VoxelNet and VirConvNet encode the LiDAR and virtual points features, respectively. Next, RoIs are generated and refined by the backbone features under different transformations. At last, the refined RoIs from different stages are fused by boxes voting~\cite{CasA}.}
   \label{fig:virconvt}
\end{figure}

\begin{table*}[!th]
	\centering
    \resizebox{\textwidth}{!}{
	\begin{tabular}{c | c | c | c c c  | c c c | c  }
    \hline
		\multirow{2}*{Method}& \multirow{2}*{Reference} &\multirow{2}*{Modality} &\multicolumn{3}{c|}{Car 3D AP (R40)}   &\multicolumn{3}{c|}{Car BEV AP (R40)}   &\multirow{2}*{Time (ms)}   \\
		                     &                          &                       &Easy     &Mod.    &Hard                 &Easy    &Mod.    &Hard                  &           \\
    \hline
    \hline
        %SECOND~\cite{SECOND}             &Sensors 2018              &LiDAR                  &83.34    &72.55   &65.82                &89.39   &83.77   &78.59                 &50 \\
        %PointRCNN~\cite{PointRCNN}       &CVPR 2019              	&LiDAR	                &86.96	  &75.64   &70.70                &92.13	  &87.39   &82.72	              &100 \\
        PV-RCNN~\cite{PV-RCNN}	         &CVPR 2020	                &LiDAR                	&90.25	  &81.43   &76.82	             &94.98	  &90.65   &86.14                 &80* \\
        Voxel-RCNN~\cite{Voxel-RCNN}	 &AAAI 2021                  &LiDAR	                &90.90    &81.62   &77.06	             &94.85	  &88.83   &86.13                 &40  \\
        CT3D~\cite{CT3D}   	             &ICCV 2021                  &LiDAR	                &87.83	  &81.77   &77.16	             &92.36	  &88.83   &84.07            	  &70* \\
        SE-SSD~\cite{SE-SSD}             &CVPR 2021                  &LiDAR                 &91.49    &82.54   &77.15                &95.68   &91.84   &86.72	              &\textbf{30}  \\
        BtcDet~\cite{BtcDet}             &AAAI 2022                  &LiDAR	                &90.64    &82.86   &78.09                &92.81   &89.34   &84.55                 &90  \\
        CasA~\cite{CasA}	             &TGRS 2022	                 &LiDAR	                &91.58    &83.06   &80.08                &95.19   &91.54   &86.82                 &86\\
        Graph-Po~\cite{Graph-Po}	     &ECCV 2022                  &LiDAR	                &91.79    &83.18   &77.98                &95.79   &92.12   &87.11                 &60\\
    \hline
        %MV3D~\cite{MV3D}                 &CVPR 2017	                 &LiDAR+RGB             &74.97    &63.63    &54.                 &86.62   &78.93   &69.8                 &360*  \\
        F-PointNet~\cite{F-PointNet}	 &CVPR 2018	                 &LiDAR+RGB             &82.19    &69.79    &60.59               &91.17   &84.67   &74.77                 &170*  \\
        UberATG-MMF~\cite{UberATG-MMF}	 &CVPR 2019	                 &LiDAR+RGB             &88.40    &77.43    &70.22               &93.67   &88.21   &81.99                 &80  \\ 
        3D-CVF~\cite{3D-CVF}	         &ECCV 2020	                 &LiDAR+RGB             &89.20    &80.05    &73.11               &93.52   &89.56   &82.45                 &75  \\
        Focals Conv~\cite{FConv}         &CVPR 2022	                 &LiDAR+RGB             &90.55    &82.28    &77.59               &92.67   &89.00   &86.33                 &100*  \\
        VPFNet~\cite{VPFNet}             &TMM 2022 	                 &LiDAR+RGB             &91.02    &83.21    &78.20               &93.94   &90.52   &86.25                 &62  \\
        Graph-VoI~\cite{Graph-Po}        &ECCV 2022	                 &LiDAR+RGB             &91.89    &83.27    &77.78               &95.69   &90.10   &86.85                 &76  \\
        SFD~\cite{SFD}                   &CVPR 2022	                 &LiDAR+RGB             &91.73    &84.76    &77.92               &95.64   &91.85   &86.83                 &98  \\
        %TED-M~\cite{TED}                 &NeruIPS 2022	             &LiDAR+RGB             &91.61    &85.28    &80.68               &95.44   &92.05   &87.3                 &94  \\
    \hline
         VirConv-L (Ours)                &-		                     &LiDAR+RGB 	        &91.41    &85.05    &80.22               &95.53   &91.95   &87.07                 &56  \\
         VirConv-T (Ours)                &-	&LiDAR+RGB      &\textbf{92.54}&\textbf{86.25}&\textbf{81.24}        &\textbf{96.11}&\textbf{92.65}&\textbf{89.69}                 &92  \\
    \hline
    \hline
         VirConv-S (Our semi-supervised)  &-		                     &LiDAR+RGB	            &92.48    &87.20    &82.45               &95.99   &93.52   &90.38                 &92  \\

    \hline
	\end{tabular}
    }
    \caption{Car 3D detection results on the KITTI test set, where the best fully supervised methods are in bold and $*$ denotes that the result is from the KITTI leaderboard. Our VirConv-T outperforms all the other methods in both 3D AP and BEV AP metrics. Besides, our VirConv-L runs fast at 56ms with 85.05 AP, and our VirConv-S attains a high detection performance of 87.20 AP.}
    \label{tab:kitti_test_ap}
\end{table*}

\subsection{Detection Frameworks with VirConv}
\label{sec:framework}

To demonstrate the superiority of our VirConv, we constructed VirConv-L, VirConv-T and VirConv-S from the widely used Voxel-RCNN~\cite{Voxel-RCNN} for fast, accurate and semi-supervised 3D object detection, respectively.

 \textbf{VirConv-L.}
We first construct the lightweight VirConv-L (Fig.~\ref{fig:framework} (c)) for fast multimodal 3D detection. VirConv-L adopts an early fusion scheme and replaces the backbone of Voxel-RCNN with our VirConvNet. Specifically, we denote the LiDAR points as $\mathbf{P}=\{p\}, p=[x,y,z,\alpha]$, where $x,y,z$ denotes the coordinates and  $\alpha$ refers intensity. We denote the virtual points as $\mathbf{V}=\{v\}, v=[x,y,z]$. We fuse them into a single point cloud $\mathbf{P}^*=\{p^*\}_k,p^*=[x,y,z,\alpha,\beta]$, where $\beta$ is an indicator denoting where the point came from. The intensity of virtual points is padded by zero. The fused points are encoded into feature volumes by our VirConvNet for 3D detection.

  \textbf{VirConv-T.}
We then construct a high-precision VirConv-T based on a Transformed Refinement Scheme (TRS) and a late fusion scheme (see Fig.~\ref{fig:virconvt}). CasA~\cite{CasA} and TED~\cite{TED} achieve high detection performance based on three-stage refinement and multiple transformation design, respectively. However, both of them require heavy computations. We fuse the two high computation detectors into a single efficient pipeline. Specifically, we first transform $\mathbf{P}$ and $\mathbf{V}$ with different rotations and reflections. Then we adopt the VoxelNet~\cite{Voxel-RCNN} and VirConvNet to encode the features of $\mathbf{P}$ and $\mathbf{V}$, respectively. Similar to TED~\cite{TED}, the convolutional weights between different transformations are shared. After that, the RoIs are generated by a Region Proposal Network (RPN)~\cite{Voxel-RCNN} and refined by the backbone features (the RoI features of $\mathbf{P}$ and $\mathbf{V}$ fused by simple concatenation) under the first transformation. The refined RoIs are further refined by the backbone features under other transformations. Next, the refined RoIs from different refinement stages are fused by boxes voting, as is done by CasA~\cite{CasA}. We finally perform a non-maximum-suppression (NMS) on the fused RoIs to obtain detection results. 

\textbf{VirConv-S.}
We also design a semi-supervised pipeline, VirConv-S, using the widely used pseudo-label method~\cite{ST3D,3DIoUMatch}. 
Specifically, first, a model is pre-trained using the labeled training data. Then, pseudo labels are generated on a larger-scale unannotated dataset using this pre-trained model. 
A high-score threshold (empirically, 0.9) is adopted to filter out low-quality labels. 
Finally, the VirConv-T model is trained using both real and pseudo labels. 
\section{Experiments}

\subsection{KITTI Datasets and Evaluation Metrics}
The KITTI 3D object detection dataset~\cite{KITTI} contains 7,481 and 7,518 LiDAR and image frames for training and testing, respectively. We divided the training data into a train split of 3712 frames and a validation split of 3769 frames following recent works~\cite{Voxel-RCNN,CasA}. We also adopted the widely used evaluation metric: 3D Average Precision (AP) under 40 recall thresholds (R40). The IoU thresholds in this metric are 0.7, 0.5, and 0.5 for car, pedestrian, and cyclist, respectively. 
We used the KITTI odometry dataset~\cite{KITTI} as the large-scale unlabeled dataset. The KITTI odometry dataset contains 43,552 LiDAR and image frames. We uniformly sampled 10,888 frames (denoted as the \textit{semi} dataset) and used them to train our VirConv-S. There is no overlap found between the KITTI 3D detection dataset and the KITTI odometry dataset after checking the mapping files released by KITTI. 

\subsection{Setup Details }
\textbf{Network details}. 
Similar to SFD, our method uses the virtual points generated by PENet~\cite{PENet}. 
VirConvNet adopts an architecture similar to the Voxel-RCNN backbone~\cite{Voxel-RCNN}. 
VirConvNet includes four levels of VirConv blocks with feature dimensions 16, 32, 64, and 64, respectively. 
The input StVD rate and layer StVD rate are set to 90\% and 15\% by default. 
On the KITTI dataset, all the detectors use the same  detection range and voxel size as CasA~\cite{CasA}.

\textbf{Losses and data augmentation}. 
VirConv-L uses the same training loss as in~\cite{Voxel-RCNN}. 
VirConv-T and VirConv-S use the same training loss as CasA~\cite{CasA}. 
In all these three pipelines, we adopted the widely used local and global data augmentation~\cite{SFD,CasA,PointRCNN}, including ground-truth sampling, local transformation (rotation and translation), and global transformation (rotation and flipping).

\textbf{Training and inference details}. 
All three detectors were trained on 8 Tesla V100 GPUs with the ADAM optimizer. We used a learning rate of 0.01 with a one-cycle learning rate strategy. We trained the VirConv-L and VirConv-T for 60 epochs. The weights of VirConv-S are initialized by the trained VirConv-T. 
We further trained the VirConv-S on the labeled and unlabeled dataset for 10 epochs. 
We used an NMS threshold of 0.8 to generate 160 object proposals with 1:1 positive and negative samples during training. During testing, we used an NMS threshold of 0.1 to remove redundant boxes after proposal refinement. 

\subsection{Main Results}

\textbf{Results on KITTI validation set.}
We report the car detection results on the KITTI validation set in Table~\ref{tab:kitti_val_ap}. Compared with the baseline detector Voxel-RCNN~\cite{Voxel-RCNN}, our VirConv-L, VirConv-T and VirConv-S show 3.42\%, 5\% and 5.68\% 3D AP(R40) improvement in the moderate car class, respectively. We also reported the performance based on the 3D AP under 11 recall thresholds (R11). Our VirConv-L, VirConv-T and VirConv-S show 2.38\%, 3.33\% and 3.54\% 3D AP(R11) improvement in the moderate car class, respectively. The performance gains are mostly derived from the VirConv design, which effectively addressed the density problem and noise problem brought by virtual points. Note that our VirConv-L also runs much faster than other multimodal detectors, thanks to our efficient StVD design.

\begin{table}[htbp]
	\centering
    \small
    %\setlength{\tabcolsep}{4.pt}
    %\resizebox{\columnwidth}{!}{
	\begin{tabular}{c |c c c |c}
		\hline
		\multirow{2}*{Method}& \multicolumn{3}{c|}{Car 3D AP (R40)} & Mod.  \\
               &Easy       &Mod.     &Hard     & AP(R11)\\
		\hline
     	\hline 
        Voxel-RCNN &92.38      &85.29        &82.86         & 84.52\\
        Voxel-RCNN(EF) &92.42      &85.78        &83.10         & 84.94\\
        Voxel-RCNN(LF) &92.91      &86.32        &83.97         & 85.23\\
        \hline 
        VirConv-L    &93.36      &88.71       &85.83        & 86.70\\
        VirConv-T     &\textbf{95.81} &\textbf{90.29}  &\textbf{88.10}        & \textbf{87.82}\\
     	\hline   
     	\hline 
        VirConv-S (semi)   &95.76      &90.97        &89.14    & 88.06\\
		\hline
	\end{tabular}
    %}
    \caption{3D car detection results on the KITTI validation set, where EF and LF denote early fusion and late fusion, respectively.}
    \label{tab:kitti_val_ap}
\end{table}

\textbf{Results on KITTI test set.}
The experimental results on the KITTI test set are reported in Table~\ref{tab:kitti_test_ap}. Our VirConv-L, VirConv-T, and VirConv-S outperform the baseline Voxel-RCNN~\cite{Voxel-RCNN} by 3.43\%, 4.63\% and 5.58\% 3D AP (R40) in the moderate car class, respectively. 
The VirConv-L, VirConv-T, and VirConv-S also outperform the best previous 3D detector SFD~\cite{SFD} by 0.29\%, 1.49\%, and 2.44\%, respectively. 
As of the date of the CVPR deadline (Nov.11, 2022), our VirConv-T and VirConv-S rank 2nd and 1st, respectively, on the KITTI 3D object detection leaderboard.
The results further demonstrate the effectiveness of our method.

\subsection{Ablation Study}
We conducted experiments on the KITTI validation set to examine the hyper-parameters and validate each component/design of the proposed method.

\begin{table}[htbp]
	\centering
    \small
    %\resizebox{\columnwidth}{!}{
	\begin{tabular}{c|c|c | c|c}
		\hline
		Setting & VirConv & TRS &3D AP & Time (ms)  \\
		\hline
        LiDAR points                   &No   & &85.29     & 38\\
		\hline
        \multirow{3}*{Virtual points}  &No   & &76.12     & 84\\
                                       &Yes   & &79.55     & \textbf{52}\\
                                       &Yes   &$\checkmark$ &\textbf{80.91}     & 71\\   
		\hline                  
        \multirow{3}*{Early fusion}    &No   & &85.78     & 88\\
                                       &Yes   & &88.71     & \textbf{56}\\
                                       &Yes   &$\checkmark$ &\textbf{88.96}     & 76\\
		\hline                      
        \multirow{3}*{Late fusion}    &No   & &86.32     & 120\\
                                       &Yes   & &88.97     & \textbf{74}\\
                                       &Yes   &$\checkmark$ &\textbf{90.29}     & 92\\
		\hline
	\end{tabular}
    %}
    \caption{Ablation results on the KITTI validation set by using different fusion scheme.  }
    \label{tab:ablation_virconv}
\end{table}
\begin{table}[htbp]
	\centering
	\small
    %\resizebox{\columnwidth}{!}{
	\begin{tabular}{c c |c c c |c }
		\hline
		with             & with      & \multicolumn{3}{c|}{ 3D AP }     &Time        \\
        StVD              & NRConv    &Easy       &Mod.     &Hard   & (ms)    \\
		\hline
         No              & No             &94.26      &87.55    &85.49   &152 \\
         Yes             & No             &94.55      &88.32    &85.95   &\textbf{87}\\
         Yes             & Yes            &\textbf{95.81}      &\textbf{90.29}    &\textbf{88.10}   &92\\
		\hline
	\end{tabular}
    %}
    \caption{Ablation results on the KITTI validation set by using different designed components.  }
    \label{tab:ablation_svd_submconv}
\end{table}

\textbf{VirConv performance with different fusion schemes.}
Virtual points only, early fusion, and late fusion are three potential choices for virtual points-based 3D object detection. To investigate the performance of VirConv under these three settings, we first constructed three baselines: Voxel-RCNN~\cite{Voxel-RCNN} with only virtual points, Voxel-RCNN~\cite{Voxel-RCNN} with early fusion, and Voxel-RCNN~\cite{Voxel-RCNN} with late fusion. 
Then we replaced the backbone of Voxel-RCNN with our VirConvNet. The experimental results on the KITTI validation set are shown in Table~\ref{tab:ablation_virconv}. 
With our VirConv, the 3D AP has significantly improved by 3.43\%, 2.93\%, and 2.65\%, under virtual points only, early fusion, and late fusion settings, respectively. Meanwhile, the efficiency significantly improves. This is because VirConv speeds up the network with the StVD design and decreases the noise impact with the NRConv design.

\textbf{Effectiveness of StVD.}
We next investigated the effectiveness of StVD. The results are shown in Table~\ref{tab:ablation_svd_submconv}. With StVD, VirConv-T not only performs more accurate 3D detection but also runs faster by about 2$\times$. The reason lies in that  StVD discards about 90\% of redundant voxels to speed up the network, and it also improves the detection robustness by simulating more sparse training samples.

\begin{figure}[t]
  \centering
   \includegraphics[width=1\linewidth]{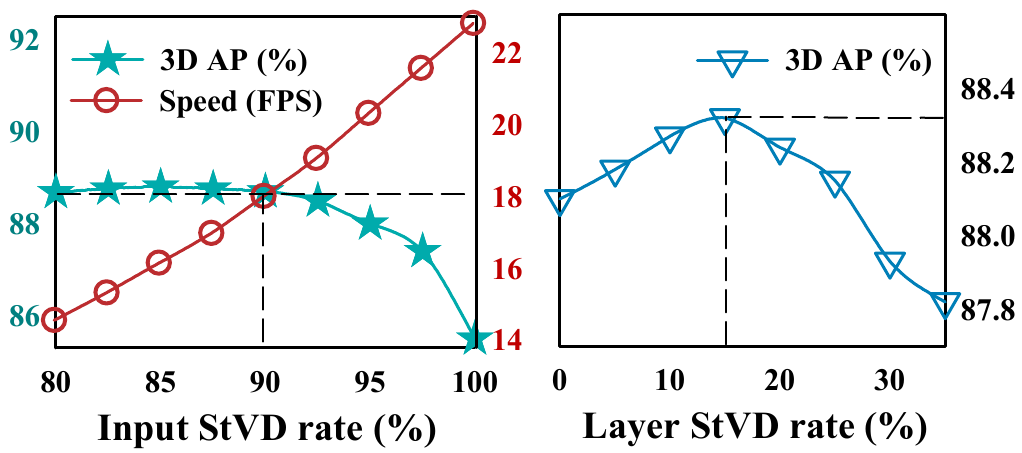}
   \caption{Left: precision and speed trade-off by using different Input StVD rate. Right: detection performance by using different layer StVD rate.}
   \label{tab:ablation_svd_rate}
\end{figure}

\textbf{Influence of StVD rate.} 
We then conducted experiments to select the best input and layer StVD rate. The results are shown in Fig.~\ref{tab:ablation_svd_rate}. We observe that using a higher input StVD rate, the detection performance will decrease dramatically due to the geometry feature loss. On the contrary, using a lower input StVD rate, the efficiency is degraded with poor AP improvement. We found that by randomly discarding 90\% of nearby voxels, we achieve the best accuracy-efficiency trad-off. Therefore, this paper adopts an input StVD rate of 90\%. Similarly, by using a 15\% layer StVD rate, we achieved the best detection accuracy.
%Thus, this paper adopts a layer StVD rate of 15\%.

\textbf{Effectiveness of NRConv.}
We then investigated the effects of NRConv using VirConv-T. The results are shown in Table~\ref{tab:ablation_svd_submconv}. With our NRConv, the car detection AP of VirConv-T improves from  88.32\% to 90.29\%. 
Since the NRConv encodes the voxel features in both 3D and 2D image space, reducing the noise impact brought by the inaccurate depth completion, the detection performance is significantly improved. 

\textbf{Effectiveness of TRS.} We conducted experiments to examine the effect of TRS in VirConv-T. The results are shown in Table~\ref{tab:ablation_virconv}. With our TRS, detectors show 1.36\%, 0.25\%, and 1.32\% performance improvement under virtual points only, early fusion, and late fusion, respectively. The performance gain is derived from the two-transform and two-stage refinement, which improves the transformation robustness and leads to better detection performance.

\begin{table}[htbp]
	\centering
	\small
    %\resizebox{\columnwidth}{!}{
	\begin{tabular}{c|c|c   c c}
		\hline 
		\multirow{2}*{Class}& \multirow{2}*{Method}& \multicolumn{3}{c}{3D AP}  \\
		                            &  & Easy & Mod. & Hard  \\
		\hline
        \multirow{2}*{Car}          & Baseline   &89.39  & 83.83     & 87.73\\
                                    &VirConv-T     &\textbf{94.98}  & \textbf{89.96}     & \textbf{88.13}\\
		\hline                  
        \multirow{2}*{Pedestrian}   &Baseline    &70.55  &62.92     &57.35\\
                                    &VirConv-T     &\textbf{73.32}  &\textbf{66.93}     &\textbf{60.38}\\
		\hline                      
        \multirow{2}*{Cyclist}      &Baseline    &89.86 &71.13     &66.67\\
                                    &VirConv-T     &\textbf{90.04} &\textbf{73.90}     & \textbf{69.06}\\
		\hline
	\end{tabular}
    %}
    \caption{3D Detection results (3D AP (R40)) of multi-class VirConv-T (KITTI validation set).  }
    \label{tab:ablation_multiclass}
\end{table}

 \textbf{Multi-class performance.}
We also trained a multi-class VirConv-T to detect car, pedestrian and cyclist class instances using a single model. 
We reported the multi-class 3D object detection performance in Table~\ref{tab:ablation_multiclass}, where the baseline refers to the multi-class Voxel-RCNN~\cite{Voxel-RCNN}. Compared with the baseline, the detection performance of VirConv-T in all classes has been significantly improved. The results demonstrate that our VirConv can be easily generalized to a multi-class model and boost the detection performance.    

\begin{figure}[t]
  \centering
   \includegraphics[width=1\linewidth]{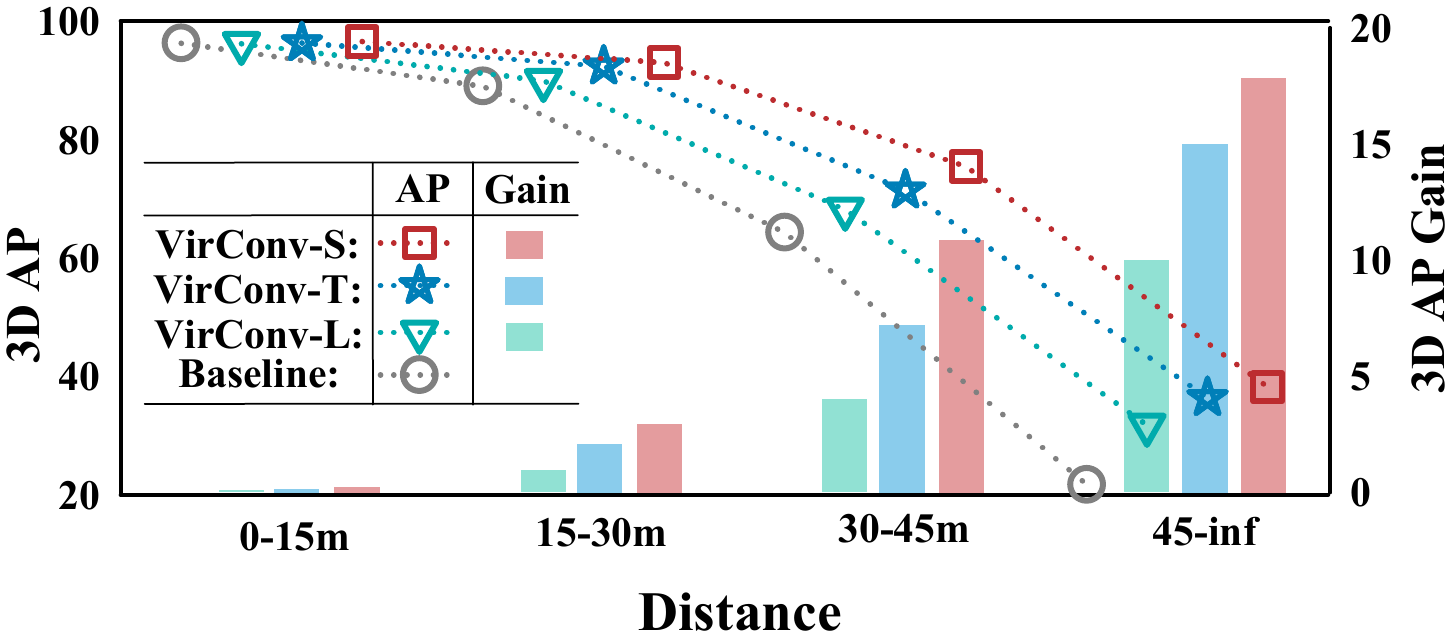}
   \caption{3D AP and performance improvement along different detection distance (KITTI validation set).}
   \label{fig:ablation_distance}
\end{figure}

 \textbf{Performance breakdown.}
To investigate where our model improves the baseline most, we evaluate the detection performance based on the different distances. The results are shown in Fig.~\ref{fig:ablation_distance}. Our three detectors have significant improvements for faraway objects because our VirConv models better geometry features of distant sparse objects from the virtual points.
 
\begin{table}[htbp]
	\centering
	\small
    %\resizebox{\columnwidth}{!}{
	\begin{tabular}{c |c c }
    \hline
Method & mAP &NDS \\
        \hline
CenterPoint + VP~\cite{MVP}& 66.4 &70.5  \\
CenterPoint + VP + VirConv  & \textbf{67.2} &\textbf{71.2} \\
        \hline
TransFusion~\cite{TransFusion} &\textbf{68.9} &71.7 \\
TransFusion-L+VP  & 66.7 &70.8 \\
TransFusion-L +VP + VirConv & 68.7 &\textbf{72.3} \\
        \hline
	\end{tabular}
    %}
    \caption{3D detection results on the nuScenes test set. }
    \label{tab:ablation_nus}
\end{table}

 \textbf{Evaluation on nuScenes test set.}
To demonstrate the universality of our method, we conducted an experiment on the nuScenes~\cite{nuScenes} dataset. we compared our method with CenterPoint + VP (virtual point), TransFuison-L + VP and TransFusion. We adopted the same data augmentation strategy as TransFuison-L and trained the network for 30 epochs on 8 Tesla V100 GPUs. The results on the nuScenes test set are shown in Table~\ref{tab:ablation_nus}. With VirConv, the detection performance of CenterPoint + VP and TransFuison-L + VP has been significantly improved. In addition, the TransFusion-L with VirConv even surpasses the TransFusion in terms of NDS, demonstrating that our model is able to boost the virtual point-based detector significantly. 

\section{Conclusion}
This paper presented a new VirConv operator for virtual-point-based multimodal 3D object detection. VirConv addressed the density and noise problems of virtual points through the newly designed Stochastic Voxel Discard and Noise-Resistant Submanifold Convolution mechanisms. Built upon VirConv, we presented VirConv-L, VirConv-T, and VirConv-S for efficient, accurate, and semi-supervised 3D detection, respectively. Our VirConvNet holds the leading entry on both KITTI car 3D object detection and BEV detection leaderboards, demonstrating the effectiveness of our method. 

\textbf{Acknowledgements }
This work was supported in part by the National Natural Science Foundation of China (No.62171393), and the Fundamental Research Funds for the Central Universities (No.20720220064).

%%%%%%%%% REFERENCES
{\small
\bibliographystyle{ieee_fullname}
\bibliography{egbib}
}

\end{document}